%% file: main.tex
\documentclass[nohyperref]{article}
\usepackage{graphicx}

\usepackage{microtype,graphicx,subfigure,booktabs,multirow,enumitem}
\usepackage{hyperref}

\usepackage[accepted, nohyperref]{icml2023}

\usepackage{amsmath,amssymb,mathtools,amsthm}

\usepackage[capitalize,noabbrev]{cleveref}

\theoremstyle{plain}

\theoremstyle{definition}

\theoremstyle{remark}

\usepackage{pifont}
\newcommand{\cmark}{\ding{51}}%
\newcommand{\xmark}{\ding{55}}%

\usepackage[textsize=tiny]{todonotes}

\definecolor{BrickRed}{rgb}{0.6,0,0}
\definecolor{RoyalBlue}{rgb}{0,0,0.8}
\definecolor{Tdgreen}{rgb}{0,0.4,0.7}
\definecolor{pinegreen}{rgb}{0.0, 0.47, 0.44}
\definecolor{cornellred}{rgb}{0.7, 0.11, 0.11}
\definecolor{cadmiumgreen}{rgb}{0.0, 0.42, 0.24}
\definecolor{spirodiscoball}{rgb}{0.06, 0.75, 0.99}

\icmltitlerunning{Debiased Distillation by Transplanting the Last Layer}

\begin{document}

\twocolumn[
\icmltitle{Debiased Distillation by Transplanting the Last Layer}

\icmlsetsymbol{equal}{*}

\begin{icmlauthorlist}
\icmlauthor{Jiwoon Lee}{ee}
\icmlauthor{Jaeho Lee}{ee,ai}
\end{icmlauthorlist}

\icmlaffiliation{ee}{Department of Electrical Engineering, POSTECH}
\icmlaffiliation{ai}{Graduate School of Artificial Intelligence, POSTECH}

\icmlcorrespondingauthor{Jaeho Lee}{jaeho.lee@postech.ac.kr}

\icmlkeywords{Debiasing, Knowledge Distillation, Spurious Correlation}

\vskip 0.3in
]
\printAffiliationsAndNotice{}

\begin{abstract}
\input{contents/abstract}
\end{abstract}

\input{contents/introduction}

\input{contents/related}

\input{contents/setup}

\input{contents/method_new}

\input{contents/experiments}

\input{contents/conclusion}


\bibliography{example_paper}
\bibliographystyle{icml2023}

\newpage
\appendix
\onecolumn
\input{contents/appendix/expdetail}

\input{contents/appendix/_Appd_baseline_projector}
\input{contents/appendix/_tsne}
\input{contents/appendix/different_teacher}

\end{document}

%% file: contents/abstract.tex
Deep models are susceptible to learning spurious correlations, even during the post-processing. We take a closer look at the knowledge distillation---a popular post-processing technique for model compression---and find that distilling with biased training data gives rise to a biased student, even when the teacher is debiased. To address this issue, we propose a simple knowledge distillation algorithm, coined DeTT (Debiasing by Teacher Transplanting). Inspired by a recent observation that the last neural net layer plays an overwhelmingly important role in debiasing, DeTT directly transplants the teacher's last layer to the student. Remaining layers are distilled by matching the feature map outputs of the student and the teacher, where the samples are reweighted to mitigate the dataset bias. Importantly, DeTT does not rely on the availability of extensive annotations on the bias-related attribute, which is typically not available during the post-processing phase. Throughout our experiments, DeTT successfully debiases the student model, consistently outperforming the baselines in terms of the worst-group accuracy.


%% file: contents/introduction.tex
\section{Introduction}
While recent applications of deep learning have demonstrated its capability to capture the core knowledge in the training data, it has also been observed that neural networks are prone to learning undesirable correlations that are easier to learn than the desired ones \citep{geirhos2020}. In many training datasets, there exist attributes which are \textit{spuriously correlated}---i.e., misleadingly correlated but not causally related---to the label. For instance, a dataset of birds may contain numerous images of birds on diverse background landscapes. It is likely that the vast majority of the images of waterbirds (the label) appear on images with the water background (spuriously correlated attribute), but not necessarily so, e.g., ``a duck on the grass.'' Given such training datasets, neural networks often learn to make predictions based on the backgrounds instead of the birds. This problem has gained much attention, and a wealth of \textit{debiasing} algorithms has been proposed \citep{sagawa2019distributionally,nam2020learning,liu2021just}. These algorithms address the spurious correlation by maximizing the \textit{worst-group accuracy} among all subgroups defined by the label and spurious correlated attribute pairs; if a model can classify a duck on the grass and the water equally well, it may be viewed as not relying on the spurious correlation.

\input{resource/fig/method}

After debiased training, however, neural network models often need to go through several post-processing steps before being deployed to the end users \citep{huyen}. Such post-processing opens up an additional window of vulnerability to bias. One of the most widely used post-processing routines is the \textit{model compression}, such as pruning or quantization \citep{han16}. It has been discovered in recent works that standard compression algorithms often disproportionately impact the performance of each subgroup, resulting in a highly biased model \citep{hooker2020characterising,lukasik2021teacher}. Following this observation, several bias-aware compression methods have been proposed to train debiased lightweight models \citep{simonsays,lin2022fairgrape}.

This paper focuses on identifying and resolving the interplay of the spurious correlation with the \textit{knowledge distillation} (KD), i.e., regularizing the training of lightweight student models with larger-scale teacher models \citep{hinton2015distilling}. KD is a widely used post-processing technique, both as a standalone model compression method and as a post-compression subroutine applied after pruning or quantization, e.g., TernaryBERT \citep{ternarybert}. Under a related setup of learning with long-tailed data \citep{van2017devil}, prior works have shown that KD brings highly disproportionate performance gains (or sometimes even degradations) to different ImageNet classes \citep{lukasik2021teacher,wang2022robust}. None of these, however, studies the effect of KD on spuriously correlated dataset; the setup critically differs from the long-tail literature in the sense that it is concerned with shortcut learning which does not necessarily originate from class imbalances. Furthermore, the target of classification is much more decoupled from the subgroup identity in the spurious correlation literature than in the long-tail learning.

\textbf{Contribution.} In this paper, we consider the task of distilling the knowledge from a debiased teacher, and find that the vanilla KD often trains a \textit{biased student} whenever the training dataset contains a spurious correlation. For example, vanilla KD on the MultiNLI dataset \citep{williams2017broad} results in over $30\%\mathrm{p}$ degradation in the worst-group performance ($78.8\% \to 46.2\%$). Importantly, the worst-group performance of the student is \textit{worse} than the student debiasedly trained without any distillation, making it questionable whether the standard KD algorithm can effectively transfer the teacher knowledge on debiasing.

To address this problem, we propose an algorithm for distilling the knowledge from debiased teachers, coined DeTT (debiasing by teacher transplanting; \cref{fig:sub1}). In a nutshell, DeTT consists of two steps: 
\begin{itemize}[leftmargin=*,topsep=0pt,parsep=-1pt]
\item \textit{Transplanting:} DeTT directly transplants the most bias-critical layer of the teacher to the student---the last layer. This design is inspired by recent observations suggesting that the last layer of the model plays an overwhelmingly important role in debiasing; the (oracle) last layer retraining of a vanilla ERM model achieves competitive performance to state-of-the-art debiasing methods \citep{kirichenko2022last}, while the retraining does not improve an already-debiased model \citep{izmailov2022feature}.
\item \textit{Feature Distillation:} We match the feature map outputs of the teacher and student models with an MSE loss. Importantly, we upweight the samples that are identified as bias-conflicting samples, where we use a separately trained biased model for the identification. We find that such upweighting significantly boosts the worst-group performance. This suggests that the feature map may also play a nonnegligible substantial role in debiasing a neural network, contrary to the prior belief.
\end{itemize}
We highlight that DeTT utilizes the labels only for identifying the samples to upweight to debiasedly train the feature map. This property leads to two practical advantages of the method: (1) DeTT can be applied to scenarios where we only have a limited number of samples with annotations on the spuriously correlated attributes, as in \citet{nam2020learning,liu2021just}, during the distillation phase. (2) DeTT can alternatively utilize external unlabeled dataset from other domains, which presumably contains less spuriously correlated attribute for the task, for the feature distillation. Indeed, we find that DeTT using ImageNet for distillation outperforms several distillation algorithms, giving a strong baseline that does not use any type of the bias supervision except for the teacher model itself.

Throughout our experiments on four vision and language debiasing benchmarks, DeTT consistently outperforms the baseline knowledge distillation and debiased training algorithms (summarized in \cref{fig:sub2}). Here, we observe that \textit{upweighting} is essential in the feature map distillation for a good debiasing performance; in Waterbirds dataset, it gives more than $16\%\mathrm{p}$ gain in the worst-group accuracy.

We believe that proposed DeTT may be a strong baseline for the future debiasing algorithms for distillation.

%% file: resource/fig/method.tex
\begin{figure*}[t]
  \centering
    \subfigure[A visual illustration of how DeTT works]{
      \includegraphics[width=0.64\textwidth]{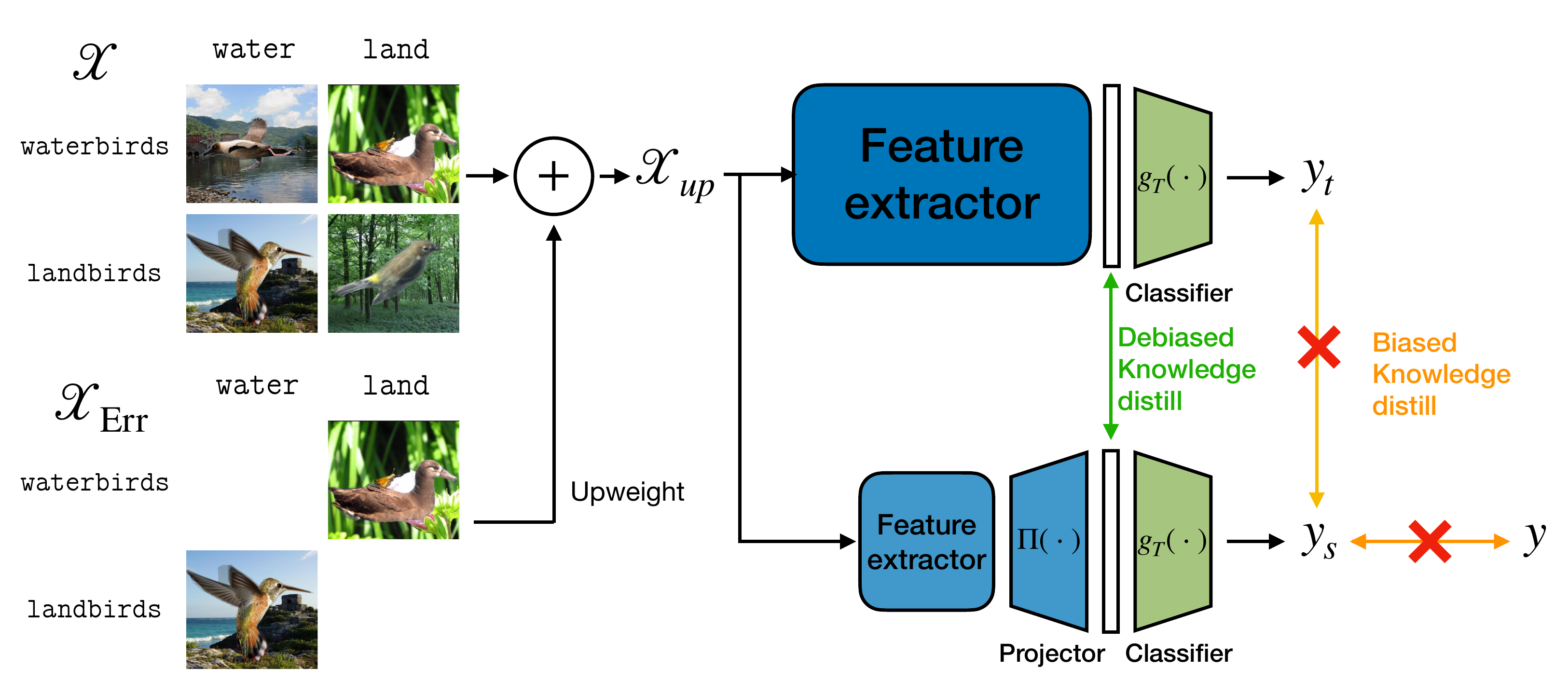}
      \label{fig:sub1}
    }
    \subfigure[Comparison of worst-group accuracies]{
      \includegraphics[width=0.3\textwidth]{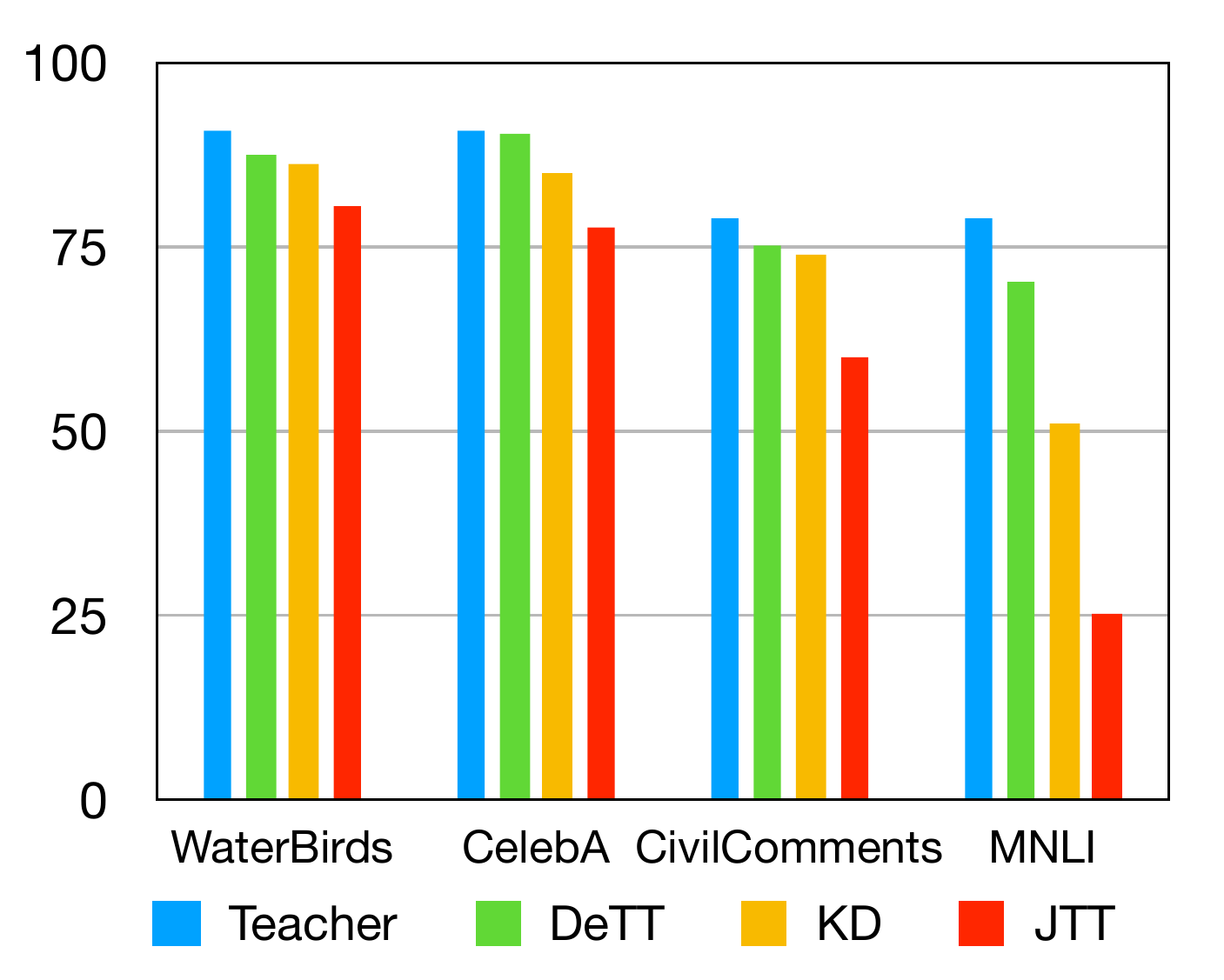}
      \label{fig:sub2}
    }
    
    \caption{A brief overview of the proposed DeTT. (a) A visual illustration of how DeTT works: We use a separately trained biased model to select and upweight the bias-conflicting samples $\mathcal{X}_{\mathrm{Err}}$. Then, we distill the feature map with the upweighted dataset, where we attach an additional projector to match the teacher and student dimensions. For the last layer, we transplant the last layer of the teacher directly.
    (b) Comparison of worst-group accuracies: DeTT achieves the best accuracy among all compared student models, achieving a worst-group accuracy that closely matches the teacher's performance.
    }
    \label{fig:method}
\end{figure*}

%% file: contents/related.tex
\section{Related Work}
In this section, we briefly overview existing work on debiasing and the knowledge distillation.

\textbf{Debiasing.} The potential harms of spurious correlations residing in the training dataset have received a lot of attention recently. The literature is distinguished from the field of ``learning from class-imbalanced data,'' in the sense that its main cause is attributed to the easy-to-learn nature of spuriously correlated attributes, instead of the imbalance in subgroup population \citep{geirhos2020}.

Many \textit{debiasing} strategies to prevent fitting the spurious correlation has been proposed.  \citet{sagawa2019distributionally} formulates debiasing as a worst-group loss minimization, leveraging the fact that if one achieves a good performance on the subgroup of bias-conflicting samples, then the model can be viewed as not utilizing the spurious correlation. Subsequent works mostly focus on a more challenging setup of debiasing without annotations on the spuriously correlated attribute, inspired by the fact that the bias in a dataset is difficult to detect \citep{nam2020learning,sohoni20,liu2021just}. A popular line of works proposes debiasing algorithms based on sample reweighting \citep{nam2020learning,liu2021just,idrissi2022simple}, while a more recent group of works focus on the role that the last layer plays in debiasing \citep{kirichenko2022last,izmailov2022feature}. Our algorithm is inspired by both lines of work, utilizing resampling and paying a special attention to the last layer.

\textbf{Knowledge distillation (KD).} 
Knowledge distillation \citep{hinton2015distilling} is a model compression technique that uses the predictions of a large-scale teacher models to enhance the training of small student models. KD also plays a significant role outside model compression, showing its power to help training of a similar-sized model \citep{furlanello2018born}. Apart from using the teacher predictions, several other frameworks have been proposed, including the feature distillation \citep{romero2014fitnets} or distilling the inter-sample relations \citep{park19}. 
Recently, \citet{chen2022knowledge} distills the feature map with the mean square loss and reuses part of teacher to perfectly mimic the teacher action; this work is most closely related to our method, which also reuses the last layer and distills the feature map.

\textbf{Knowledge distillation from imbalanced dataset.}
Several prior works study knowledge distillation under the presence of class imbalance, which is a closely related field to debiasing.
\citet{lukasik2021teacher} shows that KD algorithms can distill not only the `good knowledge' of the teacher, but also the bad knowledge of the teacher; student networks are likely to boost their performance by reproducing the teacher's behavior in an exaggerated way. This gives rise to the network which is more correct on the samples teacher gets correct, and more wrong on the samples teacher gets wrong. \citet{simonsays} shows that KD can help resolving the inter-class performance discrepancy resulting from pruning the model. \citet{wang2022robust} proposes a distillation algorithm that is specialized for distilling on the long-tailed dataset. These works focus primarily on the long-tail problem, which is distinct from the debiasing that we consider. To the best of our knowledge, our work is the first to consider knowledge distillation under the presence of spurious correlation in the training (or distillation) dataset.


%% file: contents/setup.tex
\section{Problem Setup}

Roughly stated, our goal is to train a small-sized \textit{debiased} classifier (student) by distilling the knowledge from a large debiased classifier (teacher), using the training dataset which may contain spurious correlations.


More concretely, we consider the following setup: Let $x \in \mathcal{X}$ be an input feature from which we want to predict the corresponding label $y(x) \in \mathcal{Y}$. In addition to the label, we assume that there exists another attribute $a(x) \in \mathcal{A}$ which may be \textit{spuriously correlated} (i.e., misleadingly correlated but not causally related) with the label. For example, the majority of the images labeled as $\mathtt{waterbird}$ may contain the background attribute $\mathtt{water}$, while there can also be images where waterbirds appear in other backgrounds (e.g., ``a duck on the grass''). We say that a dataset is \textit{biased} when such spurious correlation is present in the dataset.

Given a (potentially) biased dataset, our goal is to train a small-sized student model $f_S: \mathcal{X} \to \mathcal{Y}$ which avoids making predictions based on the spuriously correlated attribute. A standard way to formalize this goal is by formulating it as a group distributionally robust optimization (group DRO) problem \citep{sagawa2019distributionally}. The group DRO framework aims to minimize the expected loss for the worst-case subgroup, where the subgroup is defined as all samples sharing the same label and the spuriously correlated attribute pair. In other words, group DRO solves
\begin{equation}
\min_{f} \max_{(y,a) \in \mathcal{Y} \times \mathcal{A}} \mathbb{E}[\ell(y,f(x))~|~y(x) = y, a(x) = a], \label{eq:gdro}
\end{equation}
where $\ell(\cdot,\cdot)$ denotes the loss function. By solving the group DRO (\cref{eq:gdro}), one can avoid learning a predictor which performs too poorly on the samples that conflict with the spurious correlation, e.g., a duck on the grass. Whenever a model is trained to (approximately) achieve this goal, we will say that the model is \textit{debiased}.

For training a debiased student model $f_S$, we assume that the learner has two ingredients at hand:
\begin{itemize}[leftmargin=*,topsep=0pt,parsep=-1pt]
\item First, we have access to a \textit{biased training dataset} containing the spurious correlation. Importantly, we assume that there is \textit{no annotation} on the spuriously correlated attribute, except for a small number of samples for validation purposes \citep{nam2020learning}. This assumption comes from the difficulty of collecting bias annotations, especially when (1) the bias is conceptually difficult to be characterized and communicated to the human labelers, or when (2) the bias attribute itself carries sensitive or private information, such as ethnicity.
\item Second, we can access a \textit{debiased teacher model} $f_T$. We assume that the teacher is large in size, so that the teacher cannot be directly used for the desired purpose (e.g., deploying on a device with limited computing resources). The teacher may have been trained either by (1) utilizing another bias-annotated dataset that is not available at the distillation stage, e.g., due to privacy reasons, or (2) using the same dataset but different debiasing algorithms which are less dependent on the availability of the bias annotations \citep{nam2020learning,liu2021just}.
\end{itemize}

As we will see in \cref{table:main}, the student distilled using a biased dataset with vanilla knowledge distillation method \citep{hinton2015distilling} suffers from a limited worst-group performance; the student often performs worse than a debiasedly trained model without any teacher.


%% file: contents/method_new.tex
\section{DeTT: Debiasing by Teacher Transplanting}

We now present DeTT (Debiasing by Teacher Transplanting), a simple yet effective algorithm for distilling the knowledge of a debiased teacher using a biased training dataset without annotations on the spuriously correlated attribute.

DeTT distills the teacher by dividing-and-conquering the last layer and the remaining layers. In other words, given a teacher model $f_T$ and the student model $f_S$, we decompose both models into two parts as
\begin{equation}
\begin{aligned}
f_T(x) &= g_T \circ \phi_T(x)\\
f_S(x) &= g_S \circ \phi_S(x),
\end{aligned}
\end{equation}
where $g_T, g_S$ denote the last fully connected layers (i.e., linear classifier) of teacher and student models, respectively, and $\phi_T, \phi_S$ denotes the preceding layers (i.e., feature extractor) of the models. Given this decomposition, we apply separate strategies to distill each component. Roughly, DeTT performs the following two operations.
\begin{itemize}[leftmargin=*,topsep=0pt,parsep=-1pt]
\item \textbf{Last layer.} We directly \textit{transplant} the last fully connected layer of the teacher model to the student model and keep the layer parameters frozen $(i.e., g_S = g_T)$. If the input dimensions of $g_T$ and the output dimension of $\phi_S$ does not match, we add additional projection layer in between (see \cref{ssec:lastlayer} for more details).
\item \textbf{Feature extractor.} We train the student model feature extractor $\phi_S$ by distilling the feature map outputs from the teacher feature extractor $\phi_T$, using the biased training data. Importantly, we upweight the data samples that have been misclassified by another separately trained \textit{identification model} which has the same network architecture as the student model; we find this procedure essential for boosting the worst-group performance (see \cref{ssec:restlayers} for details).
\end{itemize}

\textbf{Intuition.} This separate processing of the last layer and preceding layers is inspired by a recent series of observations suggesting that the last layer may play an overwhelmingly important role in model debiasing than the rest of the layers; \citet{kirichenko2022last} observe that one can effectively debias a classifier trained by vanilla ERM (empirical risk minimization) procedure by simply retraining its last layer, while \citet{izmailov2022feature} report that the last layer retraining does not improve the worst-group performance of the model trained by group DRO. The latter observation, in particular, suggests that the last layer of a group DRO model (or potentially any debiasedly trained model) may contain rich information for a debiased classification. This motivates us to design a \textit{last-layer-centric} distillation algorithm, DeTT.

We note that the ideas of \textit{transplanting the last layer} and the \textit{feature distillation} of the teacher model are not entirely new. The feature distillation has been first proposed by \citet{romero2014fitnets} and has been revisited recently by \citet{yang21}. The effectiveness of transplanting the last layer with distilled features has been recently discovered by \citet{chen2022knowledge}, which is the most closely related work to ours. Our work contributes to this line of work by (1) identifying that such method gives a reasonable starting point for distilling debiased classifiers, and (2) adopting sample upweighting for resolving the feature-level bias, which gives a significant boost to the worst-group performance of the student.

We also note that, in DeTT, the labels in the training dataset are used only for identifying the samples to upweight, by which we mitigate the dataset bias. In other words, if we have an access to an external unlabeled training dataset (e.g., unlabeled ImageNet) that does not contain spurious correlation, it may be possible to debiasedly distill the feature map without any upweighting. We test this idea in section \ref{ssec:ood}, where we find that unsupervised distillation indeed leads to a competitive worst-group performance of the student.

In the remainder of this section, we describe each distillation procedure with more technical details.

\subsection{Last Layer: Transplanting the Teacher Classifier}\label{ssec:lastlayer}
For a $k$-class classification problem, let $g_S: \mathbb{R}^{d_S} \to \mathbb{R}^k$ be the last fully connected layer of the student model, where $d_S$ is the number of last hidden layer neurons. To generate $g_S$, DeTT directly transplants the teacher linear classifier $g_T: \mathbb{R}^{d_T} \to \mathbb{R}^k$, where $d_T$ is the number of last hidden layer neurons in the teacher model.

Whenever there is no mismatch in the input dimensionality between the linear classifiers of the teacher and the student (i.e., $d_S = d_T$), we simply replace $g_S$ by $g_T$. That is, DeTT changes the student model to be
\begin{equation}
f_S(x) = g_T \circ \phi_S(x),\label{eq:transplant_student}
\end{equation}
where $\phi_S$ is the feature map of the student.

In many cases, however, we have $d_S \ne d_T$. Indeed, as we are mainly considering the student model that is smaller in size than the teacher model (so that the student can be run on a limited-resource device, whereas teacher cannot), it is likely that the student model has a smaller hidden layer dimension than the teacher. In such case, we attach a lightweight trainable projector $\Pi: \mathbb{R}^{d_S} \to \mathbb{R}^{d_T}$ in front of the transplanted $g_T$. In other words, the student is
\begin{equation}
f_S(x) = g_T \circ \Pi \circ \phi_S(x).
\end{equation}
With this student model with the transplanted last layer, we train the feature map $\Pi \circ \phi_S(x)$ by distillation loss, which we will describe in more detail in \cref{ssec:restlayers}.

\subsection{Feature Map: Distillation with Sample Upweighting}\label{ssec:restlayers}

Unlike the last layer which is directly transplanted from the teacher module to the student, the feature map of the student model is distilled from the teacher using the (biased) training data. We use the MSE loss for the distillation; in the most basic form (without any sample upweighting), DeTT trains the student feature map $\phi_S$ to solve
\begin{equation}
\min_{\phi_S} \mathbb{E}\left\|\phi_S(x) - \phi_T(x) \right\|^2,\label{eq:mseloss}
\end{equation}
where $\phi_T$ is the feature map of the teacher model and the expectation is taken with respect to the training data. Whenever there is a mismatch between the output dimensionalities of $\phi_S$ and $\phi_T$, we replace $\phi_S$ by $\Pi \circ \phi_S$ and jointly train the combined feature map, where $\Pi$ is the trainable lightweight projector (described in \cref{ssec:lastlayer}).

\textbf{Sample Upweighting.} Empirically, we observe that a na\"{i}ve distillation of the feature map (via \cref{eq:mseloss}) results in a suboptimal worst-group accuracy. Instead, one can debias the student feature map further by upweighting the samples that have been misclassified by a biasedly trained model, i.e., an \textit{identification model} \citep{nam2020learning,liu2021just}. In particular, DeTT uses the identification model proposed by \citet{liu2021just} to train a vanilla ERM model with the same size as the student, and upweight the samples misclassified by this model by the factor of $\lambda_{\mathrm{up}}$. For tuning this hyperparameter $\lambda_{\mathrm{up}}$, we use a small number of \textit{validation samples} with bias annotations, as in standard debiasing works \citep{nam2020learning,liu2021just}.

%% file: contents/experiments.tex
\section{Experiment}

In this section, we evaluate the empirical performance of the proposed method, DeTT, under four debiasing benchmarks in vision and language domains. Comparing with baseline knowledge distillation algorithms, we observe that DeTT brings a significant gain in worst-group performance.

\input{contents/experiment/1Setup}
\input{contents/experiment/2MainResult}

\input{contents/experiment/3FeatureDistill}

%% file: contents/experiment/1Setup.tex
\subsection{Experimental Setup} \label{ssec:setup}
We focus on evaluating the worst-group performance of the proposed DeTT for classification tasks.

\input{resource/table/rn50/MainTable}

\textbf{Datasets.} We use total four benchmark datasets, two from the vision domain and two from the language domain.
\begin{itemize}[leftmargin=*,topsep=0pt,parsep=-1pt]
\item \textit{Waterbirds} \citep{sagawa2019distributionally} is an artificial dataset consisting of images of birds; the images have been synthetically generated from the Caltech-UCSD Birds (CUB) dataset \citep{wah2011caltech}, by cropping an image segment of a bird and pasting it on the background image. The goal of this dataset is to classify the type of birds appearing in the image, with the label set $\{\mathtt{waterbirds}, \mathtt{landbirds}\}$. The spuriously correlated attribute is the binary set of background landscapes $\{\mathtt{water},\mathtt{land}\}$.
    
\item \textit{CelebA} \citep{liu2015deep} is a dataset consisting of images of the faces of celebrities, with annotations on $40$ different attributes of the faces. We use the blondness of the hair color $\{\mathtt{blond}, \mathtt{nonblond}\}$ as the class attribute (i.e., target of prediction). As the spuriously correlated attribute, we use the gender attribute $\{\mathtt{female},\mathtt{male}\}$.
    
\item \textit{CivilComments} \cite{koh2021wilds} is a text dataset consisting of various user-generated online comments, with annotations on the toxicity of the comment and demographic identities (e.g., gender, LGBT, ethnicity, etc.) appearing in the comment. The goal of the dataset is to classify whether a sentence is $\{\mathtt{toxic},\mathtt{nontoxic}\}$. As the spuriously correlated attribute, we use whether any of the words related to demographic identities appear in the comment $\{\mathtt{present},\mathtt{absent}\}$.

\item \textit{MultiNLI} \cite{williams2017broad} is a text dataset consisting of pairs of sentences. The task is to classify the logical relationship between two sentences, where the relationship can be $\{\mathtt{contradict},\mathtt{entail},\mathtt{neutral}\}$. The spuriously correlated attribute is the presence of the negation words (e.g., ``never''), i.e., $\{\mathtt{present},\mathtt{absent}\}$.
\end{itemize}

As previously mentioned, DeTT uses a small number of ``validation samples'' with annotations on the spuriously correlated attributes for tuning the hyperparameters. For this purpose, we use the same split as \citet{liu2021just}.

\textbf{Models.} For image datasets (Waterbirds and CelebA), we use ResNet-50 as our teacher models \citep{he2016deep}. The student models are the smaller version, ResNet-18, starting from the weights that have been pretrained on the ImageNet dataset. For the text datasets (CivilComments and MultiNLI), we use BERT as our teacher \citep{devlin19}.  The student models are the pretrained version of DistilBERT \citep{sanh2019distilbert}. Unless noted otherwise, the teacher classifiers are trained with group DRO \citep{sagawa2019distributionally}, using the group annotations in the considered datasets. For the identification models (i.e., biased models for identifying samples to upweight), we use the same architecture as the student model, without any additional projectors.

\textbf{Projectors.} For the image classification models, we use a lightweight convolutional module consisting of three convolutional layers, where the first and the last layers are 1x1 convolution and the middle layer is the 3x3 depthwise convolution with 1024 input channels and 1024 output channels. For the text classification models, we do not need any projector layers, as the dimensions match already. For a fair comparison (in model size), we also attach the projectors and use teacher-sized last layer to the baseline student that does not use any layer transplanting.

\textbf{Baselines.} We compare the performance of the DeTT mainly with two baseline distillation algorithms.
\begin{itemize}[leftmargin=*,topsep=0pt,parsep=-1pt]
\item \textit{Knowledge distillation} \citep{hinton2015distilling} is the standard distillation algorithm which regularizes the student training by matching the outputs of the teacher and student models; \citet{simonsays} observes that the algorithm gives a strong baseline for mitigating the class imbalance of student models, which is (weakly) associated with the spurious correlations. The method uses the loss
\begin{equation}
\mathcal{L} = (1-\alpha) \cdot \mathcal{L}_{\mathtt{CE}} + \alpha \cdot \mathcal{L}_{\mathtt{KD}}^{(\tau)},
\label{eq:distillation}
\end{equation} where $\mathcal{L}_{\mathtt{CE}}$ is the cross-entropy loss of the student, $\mathcal{L}_{\mathtt{KD}}^{(\tau)}$ is the KL divergence between the teacher and student softmax outputs with temperature-$\tau$ scaling, and $\alpha \in [0,1]$ is the hyperparameter balancing two losses. For the temperature hyperparameter, we use $\tau = 4$, following a recent work by \citet{chen2022knowledge}. We write ``KD'' to denote this baseline method.
\item \textit{Simple KD} \citep{chen2022knowledge} is a knowledge distillation algorithm which also transplants the last layer of the teacher model to the student and distills only the feature map. Comparing with DeTT, the method lacks the upsampling step for debiasing the feature map. We write ``SimKD'' to denote this baseline.
\end{itemize}
In addition, we compare with additional baselines that does not perform any knowledge distillation. In particular, we compare with two debiasing algorithms, JTT \citep{liu2021just} and group DRO \citep{sagawa2019distributionally}, and the vanilla ERM. We note that group DRO utilizes the annotations on the spuriously correlated attribute, and thus is not intended for a direct comparison with DeTT.

\textbf{Random seeds.} All figures reported in this paper are averaged over three independent trials.

\textbf{Other experimental details.} Other experimental details, such as the optimization schedule or the choice of hyperparameters, are given in the \cref{sec:expdetails}.



%% file: resource/table/rn50/MainTable.tex
\begin{table*}[t]
    \caption{
    We compare DeTT against baseline methods on four benchmark datasets: Waterbirds, CelebA, CivilComments, and MultiNLI. For KD, we use two different levels of $\alpha$ (a hyperparameter balancing the classification and distillation loss). ``Group DRO'' denotes the student model trained via group DRO using full annotations on the spuriously correlated attributes \citep{sagawa2019distributionally}; for methods using the full annotation at the distillation phase, we mark ``annotation'' column with \cmark. \xmark\: denotes that one has access to annotations on a small number of validation samples. Best worst-group accuracies (except teacher) are marked \textbf{bold} and the second best are \underline{underlined}.
}
    \label{table:main}
        \begin{center}
            \begin{small}
                \begin{sc}
                    \resizebox{\textwidth}{!}
                    {
                    \begin{tabular}{lccccccccccr}
                    \toprule
                     \multirow{2}{*}{Method} & \multirow{2}{*}{Annotation} & \multirow{2}{*}{Teacher}& \multicolumn{2}{c}{Waterbirds} & \multicolumn{2}{c}{CelebA} & \multicolumn{2}{c}{CivilComments} & \multicolumn{2}{c}{MultiNLI} \\
                     \cmidrule(lr{1em}){4-11}
                      &  &  & Average & Worst & Average & Worst & Average & Worst & Average & Worst\\
                      \midrule
                      Teacher & \cmark & \xmark & 91.9 & 90.8 & 92.9 & 90.7 & 85.0 & 78.9 & 80.3 & 78.8\\
                    \midrule
                     \multirow{2}{*}{group DRO} & \multirow{2}{*}{\cmark} & \multirow{2}{*}{\xmark} & 87.0 & \underline{81.1} & 93.0 & 86.4 & 82.2 & \textbf{77.3} & 50.7 & 47.5\\
                    & &  & \tiny ($\pm 1.65$) & \tiny ($\pm 1.60$) & \tiny ($\pm 0.31$) & \tiny ($\pm 1.27$) & \tiny ($\pm 0.80$) & \tiny ($\pm 0.81$) & \tiny ($\pm 1.73$) & \tiny ($\pm 1.33$)\\
                    \midrule
                    \multirow{2}{*}{ERM} & \multirow{2}{*}{\xmark} & \multirow{2}{*}{\xmark} & 75.0 & 33.8 & 95.9 & 44.1 & 91.0 & 57.5 & 50.7 & 14.1 \\
                    & &  & \tiny ($\pm 1.07$) & \tiny ($\pm 3.57$) & \tiny ($\pm 0.12$) & \tiny ($\pm 1.41$) & \tiny ($\pm 0.47$) & \tiny ($\pm 5.23$) & \tiny ($\pm 1.04$) & \tiny ($\pm 5.11$)\\
                     \multirow{2}{*}{JTT} & \multirow{2}{*}{\xmark} & \multirow{2}{*}{\xmark} & 86.7 & 80.5 & 86.4 & 77.8 & 84.0 & 60.0 & 55.1 & 25.2\\
                    & &  & \tiny ($\pm 0.50$) & \tiny ($\pm 0.53$) & \tiny ($\pm 4.65$) & \tiny ($\pm 2.48$) & \tiny ($\pm 3.21$) & \tiny ($\pm 2.06$) & \tiny ($\pm 0.89$) & \tiny ($\pm 3.44$)\\
                     \multirow{2}{*}{KD($\alpha=0.5$)} & \multirow{2}{*}{\xmark} & \multirow{2}{*}{\cmark} & 87.9 & 67.9 & 95.6 & 62.0 & 87.5 & 69.1 & 57.5 & 46.2\\
                      & &  & \tiny ($\pm 0.90$) & \tiny ($\pm 0.68$) & \tiny ($\pm 0.25$) & \tiny ($\pm 4.51$) & \tiny ($\pm 2.94$) & \tiny ($\pm 6.47$) & \tiny ($\pm 0.50$) & \tiny ($\pm 0.17$)\\
                      \multirow{2}{*}{KD($\alpha=1$)} & \multirow{2}{*}{\xmark} & \multirow{2}{*}{\cmark} &
                      88.6 & 84.9 & 93.7 & 84.4 & 85.0 & 74.0 & 57.3 & 51.0\\
                     & &  & \tiny ($\pm 0.31$) & \tiny ($\pm 0.47$) & \tiny ($\pm 0.06$) & \tiny ($\pm 1.15$) & \tiny ($\pm 0.50$) & \tiny ($\pm 1.04$) & \tiny ($\pm 0.31$) & \tiny ($\pm 1.39$)\\
                    \multirow{2}{*}{SimKD} & \multirow{2}{*}{\xmark} & \multirow{2}{*}{\cmark} & 82.1 & 71.4 & 93.0 & \underline{89.0} & 87.0 & 74.0 & 70.7 & \underline{64.1} \\
                    & &  & \tiny ($\pm 1.10$) & \tiny ($\pm 2.15$) & \tiny ($\pm 0.31$) & \tiny ($\pm 0.35$) & \tiny ($\pm 0.40$) & \tiny ($\pm 2.25$) & \tiny ($\pm 0.92$) & \tiny ($\pm 1.62$) \\
                    \midrule
                    
                    \multirow{2}{*}{DeTT(ours)} & \multirow{2}{*}{\xmark} & \multirow{2}{*}{\cmark} & 90.1 & \textbf{87.5} & 92.8 & \textbf{89.5} & 86.7 & \underline{75.0} & 74.8 & \textbf{70.2} \\
                    & &  & \tiny ($\pm 0.62$) & \tiny ($\pm 1.25$) & \tiny ($\pm 0.35$) & \tiny ($\pm 0.71$) & \tiny ($\pm 0.56$) & \tiny ($\pm 2.56$) & \tiny ($\pm 0.49$) & \tiny ($\pm 0.06$) \\

                    \bottomrule
                    \end{tabular}
                    }
                \end{sc}
            \end{small}
        \end{center}
\end{table*}

%% file: contents/experiment/2MainResult.tex
\subsection{Main Result}\label{ssec:main}
The main experimental results are given in \cref{table:main}, and we make the following four observations:
\begin{itemize}[leftmargin=*,topsep=0pt,parsep=-1pt]
\item First, we observe that DeTT consistently achieves the best performance among all baselines. DeTT even outperforms the student debiased using full annotations (i.e., Group DRO), except for the CivilComments dataset.
\item Second, DeTT also tend to outperform other knowledge distillation baselines in terms of the average accuracy. In other words, the worst-group performance gain of DeTT is not from trading off the average cost, but from a better utilization of the teacher model.
\item Third, the upweighting procedure is essential for the performance. Comparing DeTT with SimKD, the performance boost is as large as $16.1\%\mathrm{p}$ (Waterbirds) or $6.1\%\mathrm{p}$. The gain is relatively small in CelebA and CivilComments, but we believe this is due to a saturation of performance; the DeTT performance is close to the teacher.
\item Fourth, the model trained only with a knowledge distillation loss (i.e., $\mathrm{KD}(\alpha=1)$) often achieves a competitive worst-group accuracy. In fact, the vanilla KD achieves a higher performance than SimKD in the waterbirds dataset, and the same performance on the CivilComments dataset.
\end{itemize}


%% file: contents/experiment/3FeatureDistill.tex
\subsection{Ablations: Transplanting and Upweighting}
\label{ssec:feat}

\input{resource/table/rn50/FeatureDistillAbla}
\input{resource/table/rn50/UpweightAbla}

We now ablate two components of DeTT: Transplanted teacher layer, and sample upweighting. In other words, we are comparing the performance of DeTT with (1) the vanilla KD using only the distillation loss, (2) the vanilla KD with upsampling, and (3) the SimKD baseline. The comparison on vision datasets is given in \cref{table:ablaFD}. We observe that both components are essential for a good worst-case performance. When the upweighting is missing, the performance dramatically fails on the Waterbirds dataset, to a level that is lower than the vanilla KD. When the transplanted teacher's last layer is missing, the performance is similar to baselines with projector even if the number of parameters decreased. Result details for baselines without projector is reported in \cref{sec:woprojector}.

We perform an additional sensitivity analysis in \cref{table:ablaUpweight}. We observe, somewhat depressingly, that upweighting does not necessarily boost the worst-group performance than the version not upweighted (SimKD). Instead, we still rely on the availability of (a small number of) validation samples with annotations on spuriously correlated attributes to find the right value of the hyperparameter $\lambda_{\mathrm{up}}$.

\subsection{Groupwise Accuracy}\label{ssec:groupwise}

We take a closer look at the subgroup accuracy of DeTT (\cref{table:distillpatternup}). One thing we note is that the worst group for the student model tends to be identical to the worst group for the teacher model, rather than the subgroup with the smallest effective number of samples (the effective number of samples means takes upweighting into account). In the Waterbirds dataset, we see that the worst-performing subgroup of DeTT is $(\mathtt{waterbirds},\mathtt{land})$, while the subgroup with the smallest effective number of samples is $(\mathtt{waterbirds},\mathtt{water})$.

\input{resource/table/rn50/distillpattern_up}
\input{resource/table/rn50/distillpattern}

We additionally look at the subgroup accuracy of SimKD (\cref{table:distillpattern}). We observe a similar pattern as in DeTT, where the worst subgroup of the student tends to be the same as the worst subgroup of the teacher model.

From these observations, we hypothesize that the transplanted last layer of the teacher to the student may allow the student to more faithfully follow the behavior of the teacher.

\subsection{DeTT with out-of-domain datasets}\label{ssec:ood}

We also test whether the DeTT can utilize out-of-domain datasets for distilling the feature map of the teacher model. In particular, we use the ImageNet dataset to distill the teachers trained on Waterbirds and CelebA, respectively. The experimental results are given in \cref{table:distillood}.

From the table, we observe that the distilled student model achieves a moderately high worst-group accuracy. In particular, in the Waterbirds dataset, the model performs better than the debiasedly trained student-size models without teachers (Group DRO and JTT), and even SimKD. In the CelebA dataset, the model performs better than JTT and the KD baseline with $\alpha=0.5$.

Considering the fact that this distillation utilizes absolutely zero supervision on the bias (even including validation samples) other than the debiased teacher, the worst-group performance of over $80\%$ accuracy provides a strong baseline.

\input{resource/table/rn50/distillood}

\subsection{Other Experiments}
We provide additional experimental results in the supplementary materials. Here, we briefly summarize what experiment they are and the takeaway messages from the results:
\begin{itemize}[leftmargin=*,topsep=0pt,parsep=-1pt]
\item \cref{sec:woprojector}: Recall that we have also attached projector layers to the baseline student models for a fair comparison. We additionally compare the empirical performance of the baselines without projectors (except for SimKD and DeTT), and find that the versions often work better than the ones without projectors. Still, DeTT outperforms these newly added baselines as well.
\item \cref{sec:tsne}: We visualize the intermediate layer outputs of the debiased teacher model via t-SNE plots to understand whether transplanting the ``last layer'' is an optimal choice. We observe that the last hidden layer indeed provides the best separation according to the target label.
\item \cref{sec:jttteacher}: Instead of the teacher trained with group DRO, we try distilling the teacher trained with JTT---a debiased training algorithm that do not require extensive annotations on the spuriously correlated attributes. Again, we find that the DeTT outperforms other baselines that does not use spurious attribute annotations for training; the only exception is on Waterbirds, where JTT (without teacher) works particularly well.

\end{itemize}

%% file: resource/table/rn50/FeatureDistillAbla.tex
\begin{table}[t]
    \caption{
    Ablations on DeTT. ``TT'' denotes the methods transplanting last layer of the teacher, and ``Up'' denotes the upsampling of bias-conflicting samples. We use $\lambda_{\mathrm{up}} = 50$, as in all other experiments. Best worst-group accuracies are marked \textbf{bold}.
    }
    \label{table:ablaFD}
        \begin{center}
            \begin{small}
                \begin{sc}
                    \resizebox{0.47\textwidth}{!}
                    {
                    \begin{tabular}{lccccccr}
                    \toprule
                     \multirow{2}{*}{Method} & \multirow{2}{*}{TT} &\multirow{2}{*}{Up}& \multicolumn{2}{c}{Waterbirds} & \multicolumn{2}{c}{CelebA}  \\
                     \cmidrule(lr{1em}){4-7}
                      &  &  & Average & Worst & Average & Worst\\
                      \midrule
                      Teacher & - & - & 91.9 & 90.8 & 92.9 & 90.7 \\
                    \midrule
                    \multirow{2}{*}{DeTT(ours)} & \multirow{2}{*}{\cmark} & \multirow{2}{*}{\cmark} & 90.1 & \textbf{87.5} & 92.8 & \textbf{89.5} \\
                     &  &  & \tiny ($\pm 0.62$) & \tiny ($\pm 1.25$) & \tiny ($\pm 0.35$) & \tiny ($\pm 0.71$) \\ 
                     \multirow{2}{*}{SimKD} & \multirow{2}{*}{\cmark} & \multirow{2}{*}{\xmark}& 82.1 & 71.4 & 93.0 & 89.0 \\
                     &  &  & \tiny ($\pm 1.10$) & \tiny ($\pm 2.15$) & \tiny ($\pm 0.31$) & \tiny ($\pm 0.35$) \\
                    \multirow{2}{*}{KD ($\alpha=1$)} & \multirow{2}{*}{\xmark} & \multirow{2}{*}{\cmark} &  88.4 & 86.3 & 93.4 & 89.1 \\
                     &  &  & \tiny ($\pm 0.50$) & \tiny ($\pm 1.23$) & \tiny ($\pm 0.06$) & \tiny ($\pm 2.17$) \\
                    \multirow{2}{*}{KD ($\alpha=1$)} & \multirow{2}{*}{\xmark} & \multirow{2}{*}{\xmark}& 88.6 & 84.9 & 93.7 & 84.4 \\
                     &  &  & \tiny ($\pm 0.31$) & \tiny ($\pm 0.47$) & \tiny ($\pm 0.06$) & \tiny ($\pm 1.15$) \\
                    \bottomrule 
                    \end{tabular}
                    }
                \end{sc}
            \end{small}

        \end{center}
\end{table}

%% file: resource/table/rn50/UpweightAbla.tex
\begin{table}[t]
    \caption{Sensitivity of the DeTT average and worst-group performance on the upweighting hyperparameter $\lambda_{\mathrm{up}}$.
    }
    \label{table:ablaUpweight}
        \begin{center}
            \begin{small}
                \begin{sc}
                    \resizebox{0.47\textwidth}{!}
                    {
                    \begin{tabular}{lcccccr}
                    \toprule
                     \multirow{2}{*}{Method} & \multirow{2}{*}{$\lambda_{\mathrm{up}}$} & \multicolumn{2}{c}{Waterbirds} & \multicolumn{2}{c}{CelebA}  \\
                     \cmidrule(lr{1em}){3-6}
                      &  & Average & Worst & Average & Worst\\
                      \midrule
                      Teacher & - & 91.9 & 90.8 & 92.9 & 90.6 \\
                    \midrule
                    \multirow{2}{*}{SimKD} & \multirow{2}{*}{0} & 82.1 & 71.4 & 93.0 & 89.0 \\
                     &  & \tiny ($\pm 1.10$) & \tiny ($\pm 2.15$) & \tiny ($\pm 0.31$) & \tiny ($\pm 0.35$)\\
                    \multirow{2}{*}{DeTT} & \multirow{2}{*}{5} & 84.5 & 76.6 & 93.5 & 88.0 \\
                     &  & \tiny ($\pm 0.80$) & \tiny ($\pm 1.47$) & \tiny ($\pm 0.15$) & \tiny ($\pm 0.86$)\\
                    \multirow{2}{*}{DeTT} & \multirow{2}{*}{20} & 88.2 & 84.7 & 93.2 & 88.3 \\
                     &  & \tiny ($\pm 0.38$) & \tiny ($\pm 0.56$) & \tiny ($\pm 0.42$) & \tiny ($\pm 1.96$)\\
                    \multirow{2}{*}{DeTT} & \multirow{2}{*}{50} & 90.1 & 87.5 & 92.8 & 89.5 \\
                     &  & \tiny ($\pm 0.62$) & \tiny ($\pm 1.25$) & \tiny ($\pm 0.35$) & \tiny ($\pm 0.71$)\\
                    \bottomrule
                    \end{tabular}
                    }
                \end{sc}
            \end{small}
        \end{center}
\end{table}

%% file: resource/table/rn50/distillpattern_up.tex
\begin{table}[t]
\caption{
Groupwise accuracy of methods using upsampling (DeTT and JTT). For DeTT, we also report the teacher accuracy. Also, we provide the effective number of samples for each subgroup after upweighting.
We note that the worst-group accuracy for Waterbirds is different from \cref{table:main}; the worst-performing group differs by the random seed.
}
\label{table:distillpatternup}
        \begin{center}
            \begin{small}
                \begin{sc}
                    \resizebox{0.47\textwidth}{!}
                    {
                    \begin{tabular}{lccc}
                    \toprule
                    \multirow{2}{*}{Waterbirds} & \# Samples & \multirow{2}{*}{JTT} & \multirow{2}{*}{DeTT(Ours)} \\ 
                     & {\small(Effective)} &  &  \\ 
                    \midrule
                    ($\mathtt{waterbirds}$,$\mathtt{water}$) & \textbf{4298} & 92.7 & 93.4 $\rightarrow$ 92.6 \tiny ($\pm 0.70$) \\
                    ($\mathtt{waterbirds}$,$\mathtt{land}$) & 6084 & \textbf{80.5} & \textbf{90.8} $\rightarrow$ \textbf{88.3} \tiny ($\pm 0.81$) \\
                    ($\mathtt{landbirds}$,$\mathtt{water}$) & 4656 & 83.2 & 91.0 $\rightarrow$ 88.8 \tiny ($\pm 1.70$) \\
                    ($\mathtt{landbirds}$,$\mathtt{land}$) & 12257 & 90.5 & 91.7 $\rightarrow$ 90.0 \tiny ($\pm 0.06$) \\
                    \midrule
                    CelebA &  &  &   \\
                    \midrule
                    ($\mathtt{nonblond}$,$\mathtt{female}$) & 200329 & 84.3 & 93.5 $\rightarrow$ 93.0 \tiny ($\pm 0.78$) \\
                    ($\mathtt{nonblond}$,$\mathtt{male}$) & 673024 & 89.7 & 93.8 $\rightarrow$ 92.3 \tiny ($\pm 0.58$) \\
                    ($\mathtt{blond}$,$\mathtt{female}$) & 466980 & 84.5 & 92.1 $\rightarrow$ 92.6 \tiny ($\pm 1.14$) \\
                    ($\mathtt{blond}$,$\mathtt{male}$) & \textbf{63837} & \textbf{80.4} & \textbf{90.7} $\rightarrow$ \textbf{89.5} \tiny ($\pm 0.71$) \\
                    \bottomrule
                    \end{tabular}
                    }
                \end{sc}
            \end{small}
        \end{center}
\end{table}

%% file: resource/table/rn50/distillpattern.tex
\begin{table}[t]
    \caption{Groupwise accuracy of methods that does not use upsampling (SimKD and ERM). For SimKD, we also report the teacher accuracy, as in \cref{table:distillpatternup}.
    }
    \label{table:distillpattern}
        \begin{center}
            \begin{small}
                \begin{sc}
                    \resizebox{0.47\textwidth}{!}
                    {
                    \begin{tabular}{lccr}
                    \toprule
                    Waterbirds & \# Samples & ERM & \multicolumn{1}{c}{SimKD} \\
                    \midrule
                    ($\mathtt{waterbirds}$,$\mathtt{water}$) & 3498 & 98.2 & 93.4 $\rightarrow$ 89.4 \tiny ($\pm 0.06$) \\
                    ($\mathtt{waterbirds}$,$\mathtt{land}$) & 184 & 58.0 & \textbf{90.8} $\rightarrow$  \textbf{71.4} \tiny ($\pm 2.15$) \\
                    ($\mathtt{landbirds}$,$\mathtt{water}$) & \textbf{56} & \textbf{33.8} & 91.0 $\rightarrow$ 84.6 \tiny ($\pm 0.79$) \\
                    ($\mathtt{landbirds}$,$\mathtt{land}$) & 1057 & 94.4 & 91.7 $\rightarrow$ 92.1 \tiny ($\pm 0.46$) \\
                    \midrule
                    CelebA &  &  &   \\
                    \midrule
                    ($\mathtt{nonblond}$,$\mathtt{female}$) & 71629 & 96.0 & 93.8 $\rightarrow$ 93.2 \tiny ($\pm 0.35$) \\
                    ($\mathtt{nonblond}$,$\mathtt{male}$) & 66874 & 99.5 & 93.8 $\rightarrow$ 93.0 \tiny ($\pm 0.46$) \\
                    ($\mathtt{blond}$,$\mathtt{female}$) & 22880 & 89.2 & 92.1 $\rightarrow$ 92.2 \tiny ($\pm 0.46$) \\
                    ($\mathtt{blond}$,$\mathtt{male}$) & \textbf{1387} & \textbf{45.6} & \textbf{90.7} $\rightarrow$  \textbf{89.0} \tiny ($\pm 0.35$) \\
                    \bottomrule
                    \end{tabular}
                    }
                \end{sc}
            \end{small}
        \end{center}
\end{table}

%% file: resource/table/rn50/distillood.tex
\begin{table}[t!]
\caption{
Groupwise accuracy of the DeTT-distilled student, when we use the ImageNet dataset for feature distillation.
}
    \label{table:distillood}
        \begin{center}
            \begin{small}
                \begin{sc}
                    \resizebox{0.47\textwidth}{!}
                    {
                    \begin{tabular}{lccr}
                    \toprule
                    Waterbirds & \# Samples & ERM & \multicolumn{1}{c}{$T \xrightarrow[]{\mathtt{ImageNet}} S$} \\
                    \midrule
                    ($\mathtt{waterbirds}$,$\mathtt{water}$) & 3498 & 98.2 & 93.4 $\rightarrow$ 85.0 \tiny ($\pm 2.81$) \\
                    ($\mathtt{waterbirds}$,$\mathtt{land}$) & 184 & 58.0 & \textbf{90.8} $\rightarrow$  \textbf{82.1} \tiny ($\pm 3.01$) \\
                    ($\mathtt{landbirds}$,$\mathtt{water}$) & \textbf{56} & \textbf{33.8} & 91.0 $\rightarrow$ 90.4 \tiny ($\pm 0.70$) \\
                    ($\mathtt{landbirds}$,$\mathtt{land}$) & 1057 & 94.4 & 91.7 $\rightarrow$ 91.1 \tiny ($\pm 0.57$) \\
                    \midrule
                    CelebA &  &  &   \\
                    \midrule
                    ($\mathtt{nonblond}$,$\mathtt{female}$) & 71629 & 96.0 & 93.8 $\rightarrow$ 92.6 \tiny ($\pm 1.19$) \\
                    ($\mathtt{nonblond}$,$\mathtt{male}$) & 66874 & 99.5 & 93.8 $\rightarrow$ 93.2 \tiny ($\pm 0.45$) \\
                    ($\mathtt{blond}$,$\mathtt{female}$) & 22880 & 89.2 & 92.1 $\rightarrow$ 90.1 \tiny ($\pm 1.28$) \\
                    ($\mathtt{blond}$,$\mathtt{male}$) & 1387 & \textbf{45.6} & \textbf{90.7} $\rightarrow$  \textbf{80.6} \tiny ($\pm 2.89$) \\
                    \bottomrule
                    \end{tabular}
                    }
                \end{sc}
            \end{small}
        \end{center}
\end{table}

%% file: contents/conclusion.tex
\section{Conclusion and discussion}
In this paper, we have discovered that knowledge distillation of a debiased teacher does not necessarily lead to a well-debiased student, whenever we train with a dataset with spurious correlations. To address this issue, we have proposed a last-layer-centric knowledge distillation algorithm, coined DeTT. Without relying on full annotations on the spurious correlated attributes, DeTT successfully debiases the student model, outperforming the baseline methods on four benchmark datasets that encompass vision and language domains; both upweighting and the last layer transplanting play key roles in boosting the worst-group performance. We also find that DeTT can also utilize out-of-domain datasets for distillation, such as ImageNet.

\textbf{Limitations and Future directions.} One of the most notable limitations of the presented DeTT is that it requires an additional \textit{projection layer} for transplanting the last layer of the student. Modern neural networks, especially the ones designed for on-device deployment, are highly optimized in terms of the size-to-performance tradeoff and the compute budget of the target device. In such cases, attaching additional layers may either undermine the peak performance or may render the modified model not deployable to the target device. In this sense, developing a knowledge distillation algorithm that does not alter the network architecture of the student model is a worthy subject of study.

Another limitation of DeTT is that the method requires separately training a biased model to select the samples to be upweighted, introducing a nonnegligible overhead in terms of the training time and computing. We believe that such overhead can be circumvented whenever we have extra supervision on the spurious correlation; how we should best utilize the extra supervision is an important future direction.

%% file: contents/appendix/expdetail.tex
\section{Experimental details}\label{sec:expdetails}

In this section, we explain further experimental details.
This section is separated into three parts: details about debiased training, details about knowledge distillation and further details about dataset.

\textbf{Debiased training.}
To train debiased teacher and baselines, we use official code released by \citet{liu2021just} and dataset introduced by \citet{sagawa2019distributionally} for our experiments.
Projector used to match dimension of classifier layer is from official code released by  \citet{chen2022knowledge}.
For NLP tasks, we use BERT and DistilBERT implemented in HuggingFace \citep{huggingface} library.
Both teacher and baseline training start from pretrained weights.\\
We search hyperparameters to train debiased model based on performance at early stopping.
Learning rate and weight decay parameter are searched for training debiased model.
Learning rate search space is divided into two types, with and without scheduler.
We tuned learning rate over $l \in \{ 1\mathrm{e}{-3},1\mathrm{e}{-4},2\mathrm{e}{-5},1\mathrm{e}{-5} \}$ for both ResNet and BERT architecture without scheduler.
In vision tasks (e.g. Waterbirds and CelebA), weight decay is searched over $\mathtt{wd} \in \{1, 0.1\}$ for strong regularization.
With scheduler, learning rate starts from 0.1 and decays when loss is on plateau for 25 and 5 epochs in Waterbirds and CelebA respectively.
In NLP tasks (e.g. CivilComments and MultiNLI), we do not use learning rate scheduler due to small number of epochs and weight decay is searched over $\mathtt{wd} \in \{0.1, 0.01, 0\}$.
Training epochs for vision task is 300 and 50 in Waterbirds and CelebA respectively.
BERT is trained with 5 epochs and DistilBERT is trained with 10 epochs.

\textbf{Distilling Knowledge to Student.}
Code for knowledge distillation (both vanilla KD and SimKD) is released by \citet{chen2022knowledge}. 
Implementation of student model is identical with that of debiased model.\\
We can divide knowledge distillation setting with two parts, end-to-end vanilla knowledge distillation and feature map distillation with transplanted layer.
For both distillation situations, we tune learning rate as described in training debiased model.
Weight decay is tuned over $\mathrm{wd} \in \{ 0.1, 0.01, 0.001\}$.
In addition, we additionally tune settings about upweighting data sample.
To select what samples to be upweighted, vanilla teacher model (i.e. model without projector) trained with ERM method is used as identification model.
Because its role is distinguishing critical samples, we select identification model from model checkpoint at 60 and 2 epoch for Waterbirds and CelebA respectively.
For NLP tasks, BERT model from checkpoint in 3 epoch is used as identification model.
Upweighting parameter is tuned over $\lambda_{\mathrm{up}} \in \{ 5,20,50\}$.
Training epochs for knowledge distillation is same with training debiased model.
When upweight is 50, training epochs in vision task is reduced to 200 and 30 for Waterbirds and CelebA, respectively. 

\textbf{Dataset.}
We describe the details of the datasets used: type of bias which exists in dataset and how much dataset is biased.
\begin{itemize}[leftmargin=*,topsep=0pt,parsep=-1pt]
\item \textit{Waterbirds} dataset is first introduced in \citet{sagawa2019distributionally} as a benchmark dataset for spurious correlation problem.
Waterbirds is artificially generated by pasting bird image to specific background.
As a result, bird type $\mathtt{\{waterbirds, landbirds\}}$ is used as class label and background feature $\mathtt{\{water, land\}}$ is used as spurious attribute.
Bias in Waterbirds is symmetric in training split (i.e. $95\%$ of $\mathtt{waterbirds}$ appear on $\mathtt{water}$ background and same ratio for $\mathtt{landbirds}$).
In test and validation split, proportion of spuriously correlated feature is balanced with given label (i.e. half is on $\mathtt{water}$ background and other help is $\mathtt{land}$ background).
It makes average accuracy to be highly affected by worst-group accuracy as represented in \cref{table:ablaFD}.

\item \textit{CelebA} dataset is a face dataset introduced in \citet{liu2015deep}.
As a benchmark for spurious correlation problem, attribute hair color $\mathtt{\{ blond, nonblond\}}$ is used as class label and gender $\mathtt{\{ female, male\}}$ is used as spurious attribute.
This new CelebA setup is used in \citet{sagawa2019distributionally}.
Different from Waterbirds dataset, its bias is unbalanced.
If label is $\mathtt{nonblond}$, proportion of $\mathtt{female}$ is $51.7\%$ (little or no bias) while proportion of $\mathtt{female}$ given $\mathtt{blond}$ label is $94.2\%$.
It means that only $(\mathtt{blond, male})$ group is victim of spurious correlation in this dataset.

\item \textit{CivilComments} dataset \citep{koh2021wilds} is a benchmark dataset used for spurious correlation problem in language model.
Classifying whether given sentence is $\mathtt{toxic, nontoxic}$ is goal of this benchmark dataset while existence of demographic identities $\mathtt{present, absent}$ (i.e. gender, LGBT, religion or race) works as spuriously correlated attribute.
In this dataset, $\mathtt{toxic}$ sentence is more likely to have demographic identities ($60.9\%$) than $\mathtt{nontoxic}$ sentence ($39.8\%$).
We can observe that CivilComments dataset is less biased among all our benchmarks.

\item \textit{MultiNLI} dataset is introduced in \citet{williams2017broad} and used as benchmark for spurious correlation problem in \citet{sagawa2019distributionally}.
This task classifies logical relationship given two sentences, whether one sentence is $\mathtt{\{ contradict, entailed, neutral\}}$ to another one.
Presence of negation $\mathtt{present, absent}$ works as spuriously correlated attribute.
If negation appears in one sentence, proportion of label $\mathtt{contradict}$ is $76.2\%$ while only $30.0\%$ of sentence pair is $\mathtt{contradict}$ when there is no negation.
In addition, it is a ternary classification task while other benchmarks are binary classification task.

\end{itemize}

%% file: contents/appendix/_Appd_baseline_projector.tex
\newpage
\section{Baseline with and without projector}\label{sec:woprojector}

\input{resource/table/rn50/Main_projector}

Recall the \cref{ssec:feat}, our experiments on \cref{table:main} is produced on ResNet-18 with projector, to match the number of parameters among baseline.
\cref{table:projector} is an expansion of \cref{table:main} to show difference between original ResNet-18 and ResNet-18 with projector.
In \cref{table:projector}, result from original ResNet-18 baseline is similar to result from ResNet-18 with projector.
We believe that it is due to initialization of projector, while original ResNet-18 training starts from ImageNet pre-trained weights.
Although some original ResNet-18 experiments work slightly better than our modified ResNet-18 baselines, DeTT still performs better than original ResNet-18 baseline.

%% file: resource/table/rn50/Main_projector.tex
\begin{table*}[h]
    \caption{
    Expansion of Table \ref{table:main} according to baselines with and without projector.
    In NLP tasks, results with projector is absent because they do not need projector for default setting.
}
    \label{table:projector}
    \vskip 0.15in
        \begin{center}
            \begin{small}
                \begin{sc}
                    \resizebox{0.8\textwidth}{!}
                    {
                    \begin{tabular}{lccccccr}
                    \toprule
                     \multirow{2}{*}{Method} & \multirow{2}{*}{Annotation} & \multirow{2}{*}{Teacher}& \multicolumn{2}{c}{Waterbirds} & \multicolumn{2}{c}{CelebA}  \\
                     \cmidrule(lr{1em}){4-7}
                      &  &  & Average & Worst & Average & Worst\\
                      \midrule
                      Teacher & \cmark & \xmark & 91.9 & 90.8 & 92.9 & 90.7\\
                    \midrule
                     \multirow{2}{*}{group DRO} & \multirow{2}{*}{\cmark} & \multirow{2}{*}{\xmark} & 87.0 & \underline{81.1} & 93.0 & 86.4\\
                    & &  & \tiny ($\pm 1.65$) & \tiny ($\pm 1.60$) & \tiny ($\pm 0.31$) & \tiny ($\pm 1.27$) \\
                     \multirow{2}{*}{$\hookrightarrow \text{w/o projector}$} & \multirow{2}{*}{\cmark} & \multirow{2}{*}{\xmark} & 87.0 & 80.0 & 92.9 & 87.6 \\
                    & &  & \tiny ($\pm 0.91$) & \tiny ($\pm 1.65$) & \tiny ($\pm 0.35$) & \tiny ($\pm 1.15$) \\
                    \midrule
                    \multirow{2}{*}{ERM} & \multirow{2}{*}{\xmark} & \multirow{2}{*}{\xmark} & 75.0 & 33.8 & 95.9 & 44.1 \\
                    & &  & \tiny ($\pm 1.07$) & \tiny ($\pm 3.57$) & \tiny ($\pm 0.12$) & \tiny ($\pm 1.41$)  \\
                    \multirow{2}{*}{$\hookrightarrow \text{w/o projector}$} & \multirow{2}{*}{\xmark} & \multirow{2}{*}{\xmark} & 71.5 & 18.6 & 96.0 & 45.6 \\
                    & &  & \tiny ($\pm 0.91$) & \tiny ($\pm 0.77$) & \tiny ($\pm 0.06$) & \tiny ($\pm 2.12$)\\
                     \multirow{2}{*}{JTT} & \multirow{2}{*}{\xmark} & \multirow{2}{*}{\xmark} & 86.7 & 80.5 & 86.4 & 77.8 \\
                    & &  & \tiny ($\pm 0.50$) & \tiny ($\pm 0.53$) & \tiny ($\pm 4.65$) & \tiny ($\pm 2.48$) \\
                    \multirow{2}{*}{$\hookrightarrow \text{w/o projector}$} & \multirow{2}{*}{\xmark} & \multirow{2}{*}{\xmark} & 86.8 & 81.8 & 89.3 & 78.3 \\
                    & &  & \tiny ($\pm 0.50$) & \tiny ($\pm 0.96$) & \tiny ($\pm 2.75$) & \tiny ($\pm 1.48$) \\

                      
                      \multirow{2}{*}{KD($\alpha=1$)} & \multirow{2}{*}{\xmark} & \multirow{2}{*}{\cmark} &
                      88.6 & 84.9 & 93.7 & 84.4 \\
                     & &  & \tiny ($\pm 0.31$) & \tiny ($\pm 0.47$) & \tiny ($\pm 0.06$) & \tiny ($\pm 1.15$) \\

                    \multirow{2}{*}{$\hookrightarrow \text{w/o projector}$} & \multirow{2}{*}{\xmark} & \multirow{2}{*}{\xmark} & 88.8 & 84.7 & 93.8 & 84.7 \\
                    & &  & \tiny ($\pm 0.36$) & \tiny ($\pm 0.42$) & \tiny ($\pm 0.35$) & \tiny ($\pm 81$)\\
                     
                    \multirow{2}{*}{SimKD} & \multirow{2}{*}{\xmark} & \multirow{2}{*}{\cmark} & 82.1 & 71.4 & 93.0 & \underline{89.0} \\
                    & &  & \tiny ($\pm 1.10$) & \tiny ($\pm 2.15$) & \tiny ($\pm 0.31$) & \tiny ($\pm 0.35$) \\
                    \midrule
                    
                    \multirow{2}{*}{DeTT(ours)} & \multirow{2}{*}{\xmark} & \multirow{2}{*}{\cmark} & 90.1 & \textbf{87.5} & 92.8 & \textbf{89.5}\\
                    & &  & \tiny ($\pm 0.62$) & \tiny ($\pm 1.25$) & \tiny ($\pm 0.35$) & \tiny ($\pm 0.71$) \\
                    \bottomrule
                    \end{tabular}
                    }
                \end{sc}
            \end{small}
        \end{center}
    \vskip -0.1in
\end{table*}

%% file: contents/appendix/_tsne.tex
\newpage
\section{Feature Level Analysis of DeTT}\label{sec:tsne}

\input{resource/fig/tsne/tsne}

In this section, we will analyze how feature map looks like in debiased model DeTT can distill robustness to student.

\cref{fig:tsne} is feature map output from models trained with Waterbirds dataset.
From \cref{sfig:t1} to \ref{sfig:tf}, feature map from debiased teacher is represented through t-SNE plot.
\cref{sfig:tf}, \ref{sfig:erm}, \ref{sfig:simkd} and \ref{sfig:dett} are feature distribution as input of teacher's classifier layer. 
In the shallowest feature map in \cref{sfig:t1}, data are classified based on background feature which is the easy-to-learn feature.
The second last feature map in \cref{sfig:t4} (while the pooled last feature map (\cref{sfig:tf}) is input of transplanted classifier) is still not aware of core feature.
However, in \cref{sfig:tf}, it shows obvious pattern to classify $\mathtt{waterbird}$ and $\mathtt{landbird}$.
Groups with same label (but different annotation) share their region in t-SNE map without overlapping different label.
Through this pattern, feature map distillation works well to distinguish labels of data even if dataset with spurious correlation is used during distillation.

With \cref{fig:tsne}, we can think about feature map distribution with biased dataset.
Bias-conflict groups (i.e. pale red and light blue data points in \cref{fig:tsne}) is more sparse than spuriously correlated groups in biased dataset.
It means that there would be inaccurate distillation on bias-conflict group's own distribution.
It makes unclear samples (i.e. samples near decision boundary) in student model, which can be observed in \cref{sfig:simkd}.
From this observation, we think that distilling fine-tuned feature extractor would reduce error for bias-conflict sample because spuriously correlated groups shares feature distribution of bias-conflict groups in \cref{sfig:tf} while \cref{sfig:erm} does not.
In other words, bias-conflict groups can distill their feature map distribution through spuriously correlated samples when feature map is fine-tuned.
Although each feature extractor with its own fine-tuned classifier can work debiased model \citep{izmailov2022feature}, fine-tuning feature extractor is recommended when considering feature distillation.

\newpage

%% file: resource/fig/tsne/tsne.tex
\begin{figure*}[h]
  \centering
    \subfigure[Feature map from teacher's first block output]{
      \includegraphics[width=0.4\textwidth]{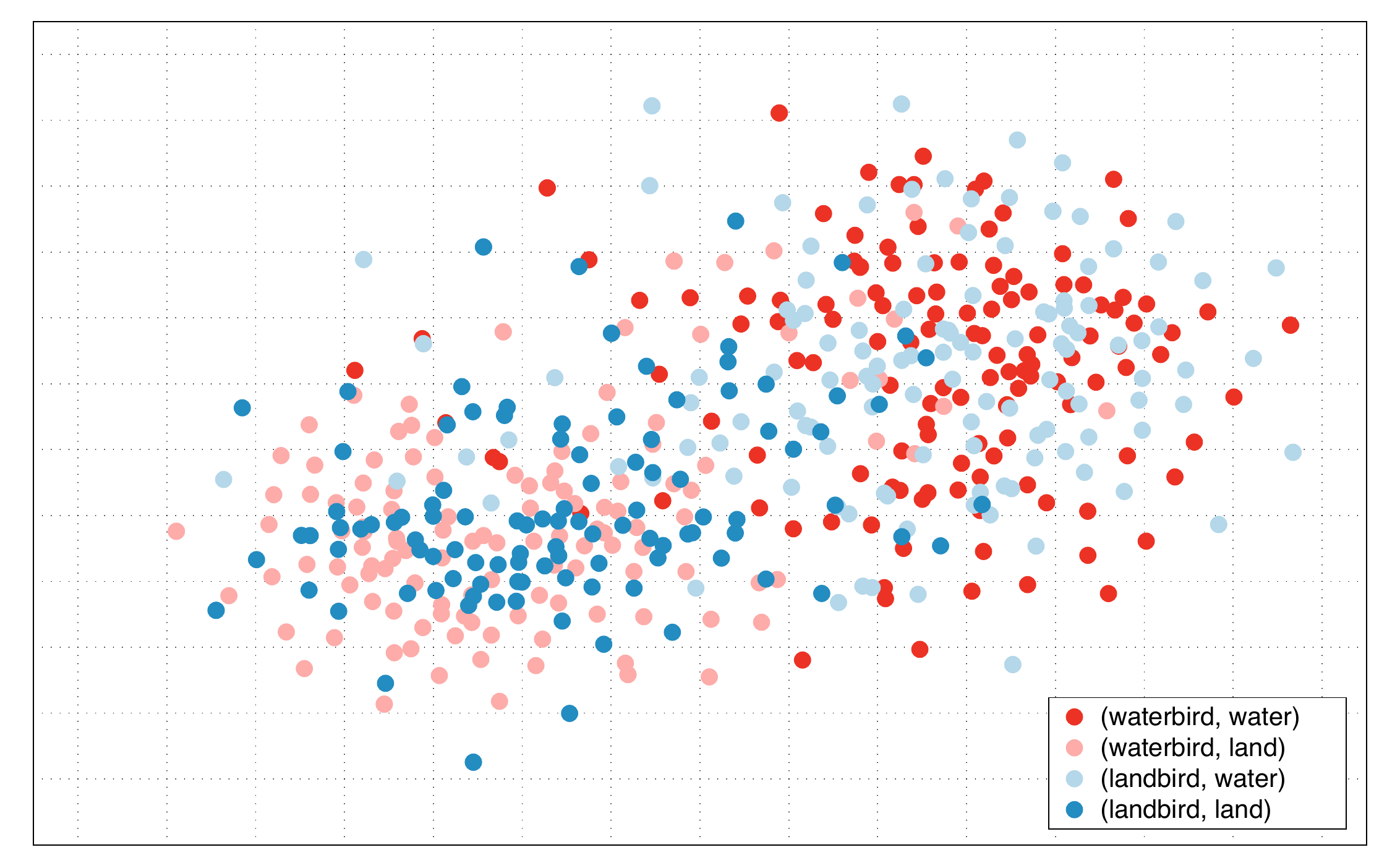}
      \label{sfig:t1}
    }
    \subfigure[Feature map from teacher's fourth block output]{
      \includegraphics[width=0.4\textwidth]{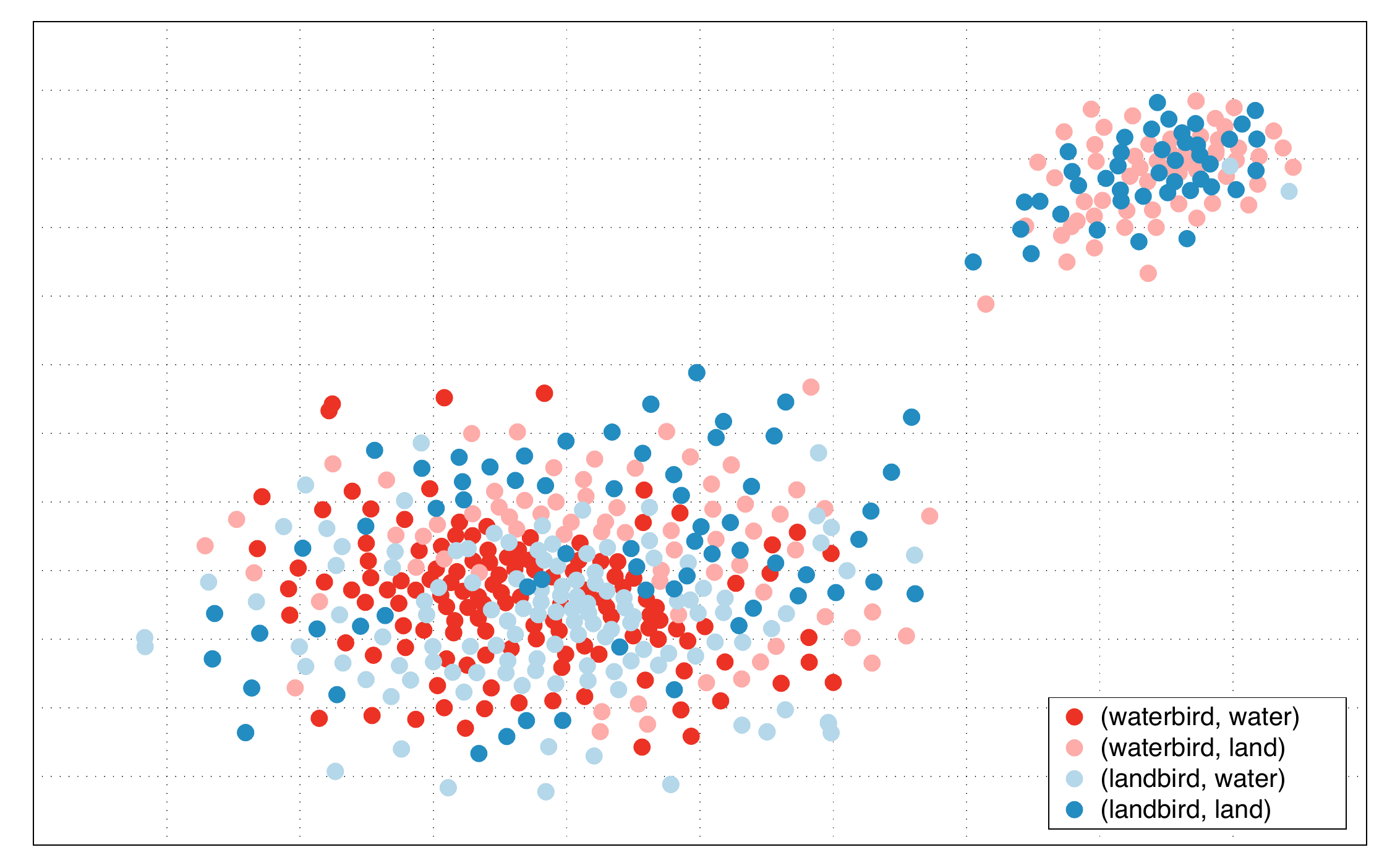}
      \label{sfig:t4}
    }
    \subfigure[Pooled feature map from teacher used in DeTT]{
      \includegraphics[width=0.4\textwidth]{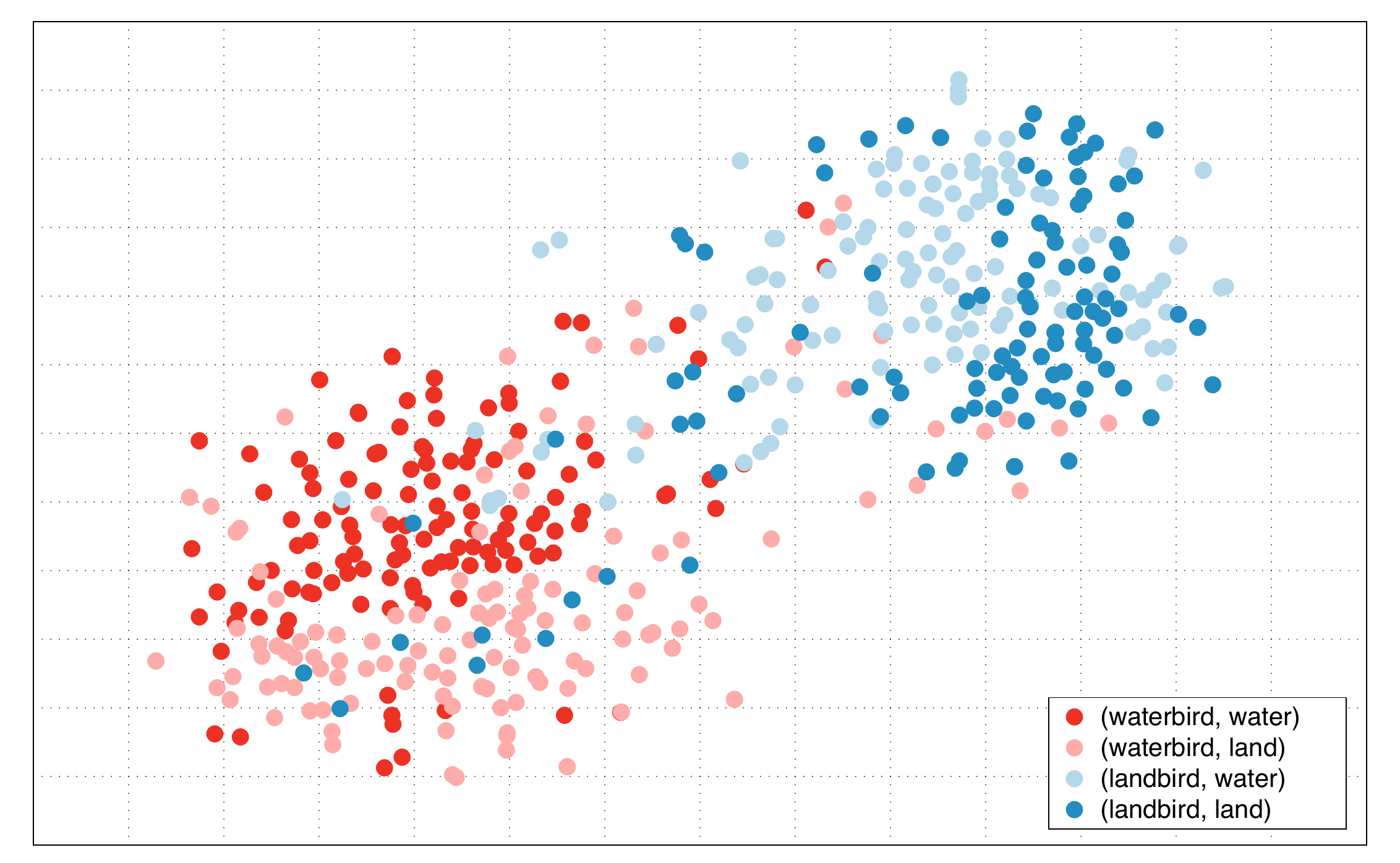}
      \label{sfig:tf}
    }
    \subfigure[Pooled feature map from student trained with ERM]{
      \includegraphics[width=0.4\textwidth]{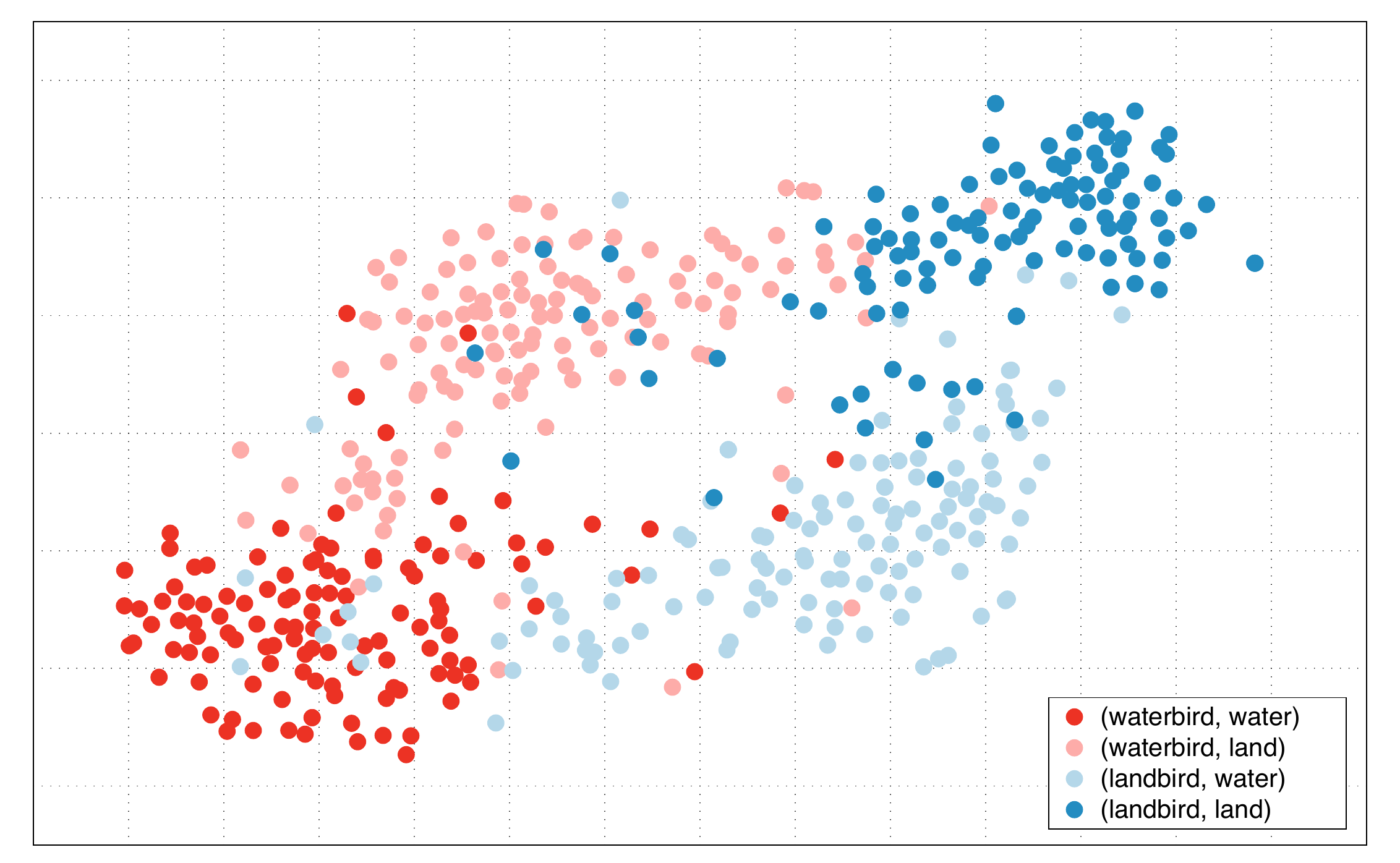}
      \label{sfig:erm}
    }
    \subfigure[Pooled feature map from student trained with SimKD]{
      \includegraphics[width=0.4\textwidth]{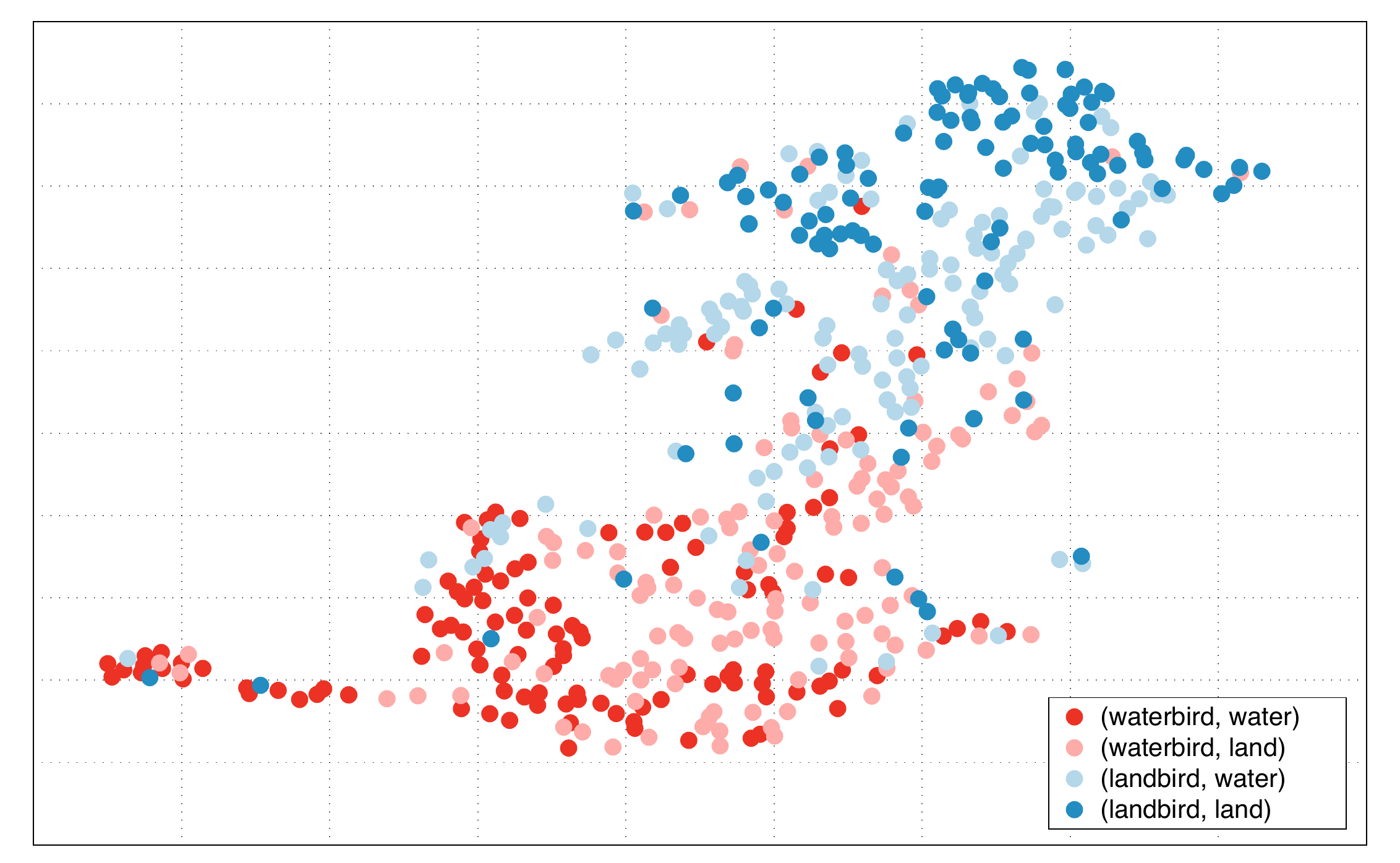}
      \label{sfig:simkd}
    }
    \subfigure[Pooled feature map from student trained with DeTT]{
      \includegraphics[width=0.4\textwidth]{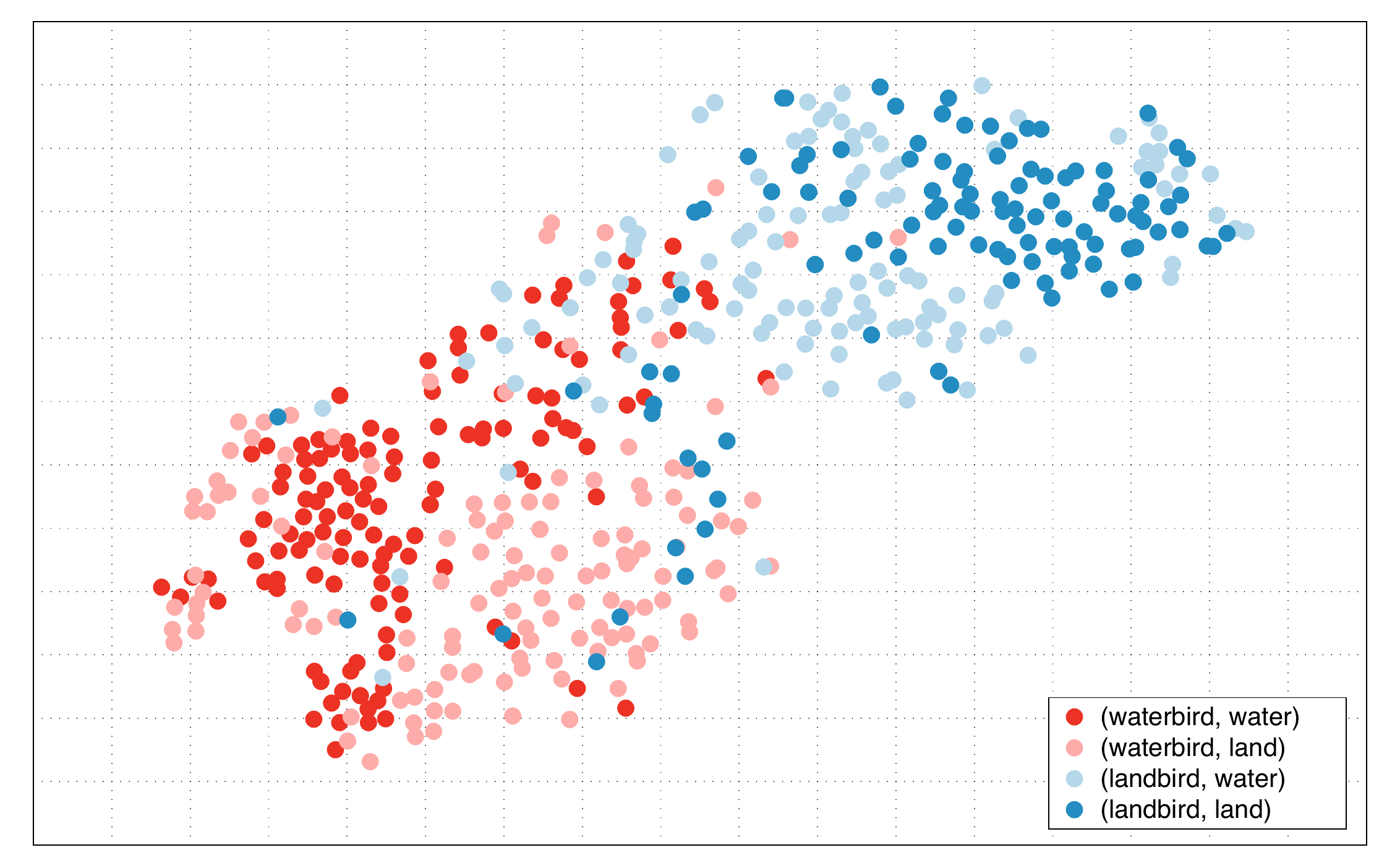}
      \label{sfig:dett}
    }
    
    \caption{
    Feature map distribution for Waterbirds dataset is represented through t-SNE method.
    From Figure \ref{sfig:t1} to \ref{sfig:tf}, feature map from teacher trained with gDRO \citep{sagawa2019distributionally} is represented.
    For Figure \ref{sfig:erm}, ResNet-18 model is trained with end-to-end ERM method and its feature map is captured.
    Figure \ref{sfig:simkd} and \ref{sfig:simkd} is feature map distribution distilled from teacher used in Figure \ref{sfig:t1} to \ref{sfig:tf}.
    }
    \label{fig:tsne}
\end{figure*}

%% file: contents/appendix/different_teacher.tex
\newpage
\section{DeTT with different teacher}\label{sec:jttteacher}
\input{resource/table/rn50/different_teacher}
To show how DeTT works for teacher with different performance, we try DeTT method using teacher trained by JTT \citep{liu2021just} method.
We find two observations from experiment in \cref{table:jttteacher}.

First observation is that performance of DeTT can drop if feature extractor is overfitted to training dataset.
In \cref{table:jttteacher}, DeTT shows the best performance among knowledge distillation baselines (i.e. KD($\alpha = 1$) and SimKD) on both Waterbirds and CelebA dataset.
However, JTT baseline outperforms DeTT in Waterbirds dataset even if JTT does not need teacher network.
We observe that training accuracy saturates to 100\% for all groups in Waterbirds during DeTT training, as teacher network does.
Student network which shows 100\% training accuracy (but has not seen any true label) means network is overfitted in feature level.
We believe this low distillation performance is due to overfitting in feature level, which does not happen in the same experiment in \cref{table:main}.
Second observation in \cref{table:jttteacher} is that DeTT shows performance gain compared with SimKD in CelebA dataset, while \cref{table:ablaUpweight} shows less performance boost after upweighting.
It supports our third belief in \cref{ssec:main}, that if student's performance does not saturate to teacher, performance can be boosted when data samples are upweighted.

%% file: resource/table/rn50/different_teacher.tex
\begin{table*}[h]
    \caption{
    Main result in \cref{table:main} is reproduced with different teacher, trained with JTT \citep{liu2021just}.
    Results form debiased training baseline (i.e. ERM, JTT \citep{liu2021just} and group DRO \citep{sagawa2019distributionally}) in this table is shared with \cref{table:main}.
}
    \label{table:jttteacher}
    \vskip 0.15in
        \begin{center}
            \begin{small}
                \begin{sc}
                    \resizebox{0.8\textwidth}{!}
                    {
                    \begin{tabular}{lccccccr}
                    \toprule
                     \multirow{2}{*}{Method} & \multirow{2}{*}{Annotation} & \multirow{2}{*}{Teacher}& \multicolumn{2}{c}{Waterbirds} & \multicolumn{2}{c}{CelebA}  \\
                     \cmidrule(lr{1em}){4-7}
                      &  &  & Average & Worst & Average & Worst\\
                      \midrule
                      Teacher & \cmark & \xmark & 92.0 & 86.0 & 94.7 & 77.2\\
                    \midrule
                     \multirow{2}{*}{group DRO} & \multirow{2}{*}{\cmark} & \multirow{2}{*}{\xmark} & 87.0 & 81.1 & 93.0 & 86.4\\
                    & &  & \tiny ($\pm 1.65$) & \tiny ($\pm 1.60$) & \tiny ($\pm 0.31$) & \tiny ($\pm 1.27$) \\
                    \multirow{2}{*}{ERM} & \multirow{2}{*}{\xmark} & \multirow{2}{*}{\xmark} & 75.0 & 33.8 & 95.9 & 44.1 \\
                    & &  & \tiny ($\pm 1.07$) & \tiny ($\pm 3.57$) & \tiny ($\pm 0.12$) & \tiny ($\pm 1.41$)  \\
                     \multirow{2}{*}{JTT} & \multirow{2}{*}{\xmark} & \multirow{2}{*}{\xmark} & 86.7 & 80.5 & 86.4 & 77.8 \\
                    & &  & \tiny ($\pm 0.50$) & \tiny ($\pm 0.53$) & \tiny ($\pm 4.65$) & \tiny ($\pm 2.48$) \\
                      \multirow{2}{*}{KD($\alpha=1$)} & \multirow{2}{*}{\xmark} & \multirow{2}{*}{\cmark} &
                      85.4 & 73.6 & 94.5 & 76.7 \\
                     & &  & \tiny ($\pm 0.06$) & \tiny ($\pm 1.60$) & \tiny ($\pm 0.83$) & \tiny ($\pm 1.05$) \\
                    \multirow{2}{*}{SimKD} & \multirow{2}{*}{\xmark} & \multirow{2}{*}{\cmark} & 78.7 & 65.2 & 94.9 & 73.9 \\
                    & &  & \tiny ($\pm 0.60$) & \tiny ($\pm 0.47$) & \tiny ($\pm 0.62$) & \tiny ($\pm 1.15$) \\
                    \midrule          
                    \multirow{2}{*}{DeTT(ours)} & \multirow{2}{*}{\xmark} & \multirow{2}{*}{\cmark} & 86.5 & 77.0 & 94.7 & 78.9\\
                    & &  & \tiny ($\pm 2.27$) & \tiny ($\pm 3.00$) & \tiny ($\pm 2.17$) & \tiny ($\pm 0.05$) \\

                    \bottomrule
                    \end{tabular}
                    }
                \end{sc}
            \end{small}
        \end{center}
    \vskip -0.1in
\end{table*}